\documentclass{article}
\usepackage[dvips]{graphicx}
\usepackage[latin1]{inputenc}
\usepackage{amssymb,amsmath,array}
\usepackage{epsfig}

\begin{document}
\title{Learning and discrimination through STDP\\
in a top-down modulated associative memory \footnote{Published in Proceedings of 14$^{th}$ European Symposium on Artifical Neural Networks (ESANN 2006), M. Verleysen (ed.), p 611-616, Bruges, Belgium} }

\author{Anthony Mouraud$^1$$^,$$^2$ and Hélène Paugam-Moisy$^1$
%
%
\vspace{.3cm}\\
%
1 - Institut des Sciences Cognitives - UMR CNRS 5015 - 69675 Bron - France
%
\vspace{.1cm}\\
2 - GRIMAAG - Université Antilles-Guyane - 97157 Pointe-à-Pitre - France
}

\maketitle

\begin{abstract}
This article underlines the learning and discrimination capabilities of a 
model of associative memory based on artificial networks of spiking neurons.
Inspired from neuropsychology and neurobiology, the model implements top-down 
modulations, as in neocortical layer V pyramidal neurons, with a learning rule based
on synaptic plasticity (STDP), for performing a multimodal association learning task. 
A temporal correlation method of analysis proves the ability of the model 
to associate specific activity patterns to different samples of stimulation. 
Even in the absence of initial learning and with continuously varying weights, 
the activity patterns become stable enough for discrimination.
\end{abstract}

\section{Introduction}
\label{intro}

In the scope of pattern recognition and machine learning, artificial neural 
networks have proved to be computationally efficient tools. However classical 
learning methods and connectionist models have shown several limitations,
e.g. fast adaptation to changing environment, or modelling temporal 
binding and synchronization phenomena for multimodal integration \cite{RD99,CK03}.

Introduced in \cite{GH92} and extensively described later \cite{GK02}, 
spiking neurons use precise timing of spike emissions as relevant neural code,
in opposition to the rate coding of the first two generations of neural networks
 \cite{Maa97}. Spike time coding increases the speed of image processing 
\cite{TDVR01} and the computational power of neural networks \cite{MM04}.
However learning in spiking neuron networks (SNNs) is not yet controlled
for performing general purpose discrimination tasks efficiently. A first track 
is to exploit neuroscientist knowledge on synaptic plasticity mechanisms, like
Spike-Timing-Dependent Plasticity (STDP) \cite{AN00,RS01}, that can be easily implemented 
in SNNs as unsupervised dynamic hebbian learning rule for on-line adaptation of weights. 

A second idea is to incorporate attentional mechanisms in the model, starting from
an architecture of bidirectional associative memory already proven efficient for
pattern recognition \cite{Rey05}. 
It is largely approved that top-down processing is involved in attention \cite{HBM00}, 
which is fundamental for efficient learning. 
From physiological studies \cite{LZS01}, we derive that neocortical layer V pyramidal 
neurons, able to integrate bottom-up and top-down signals, are well suited for modelling top-down influences in the network.

Only few work try to associate SNNs and STDP in pattern recognition tasks, 
and none of them study the influence of top-down modulations in such tasks. 
In this article we present a three layer multimodal associative memory 
coupling models of spiking pyramidal neurons and interneurons in a STDP driven 
learning and discrimination task. We show that the model is able to dynamically 
associate specific patterns of activity to bimodal stimulations and we study 
the influence of top-down modulations on learning speed and discrimination.

\section{Multimodal bidirectional associative memory}
\label{model}

\begin{figure}[hbt]
\begin{tabular}{c|c}
\includegraphics[scale=0.55]{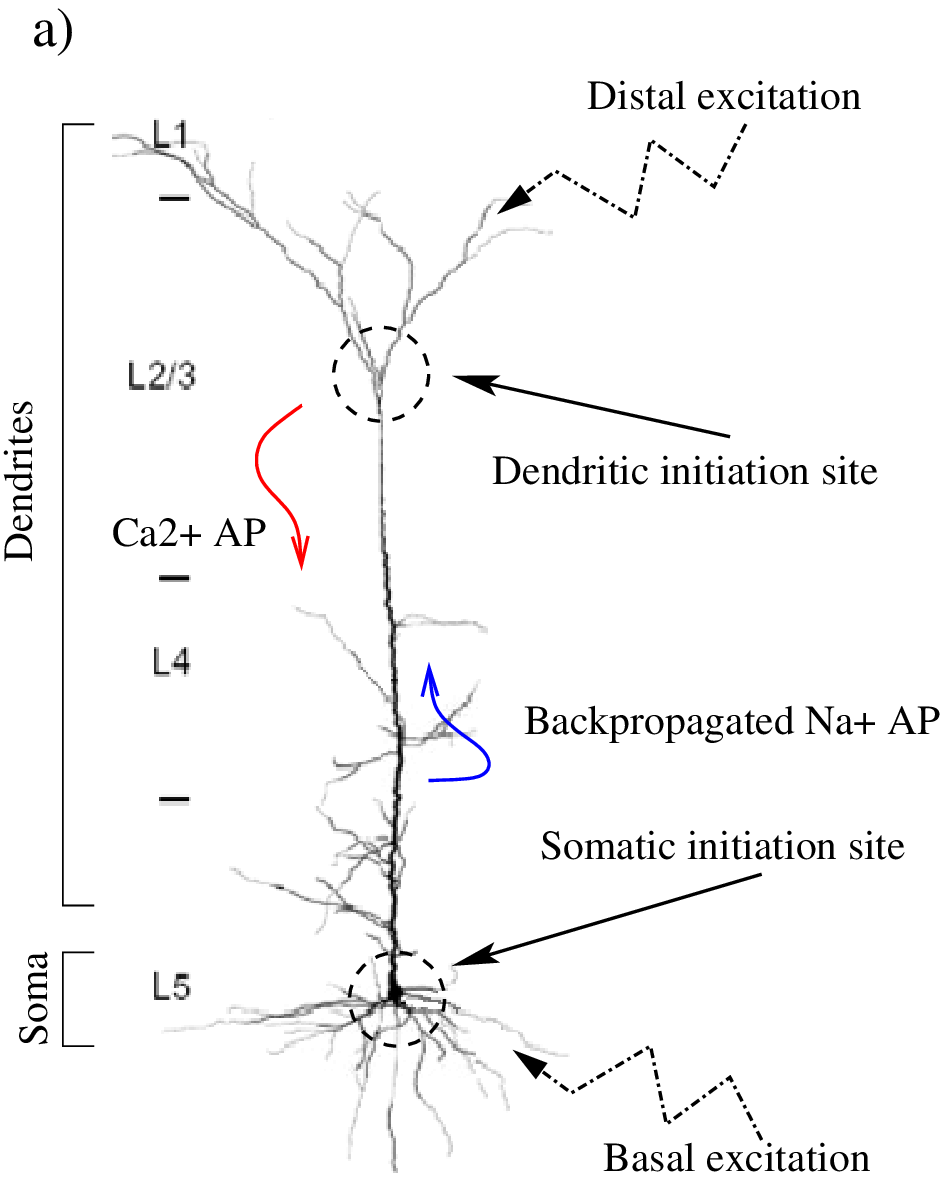}
&
\includegraphics[scale=0.43]{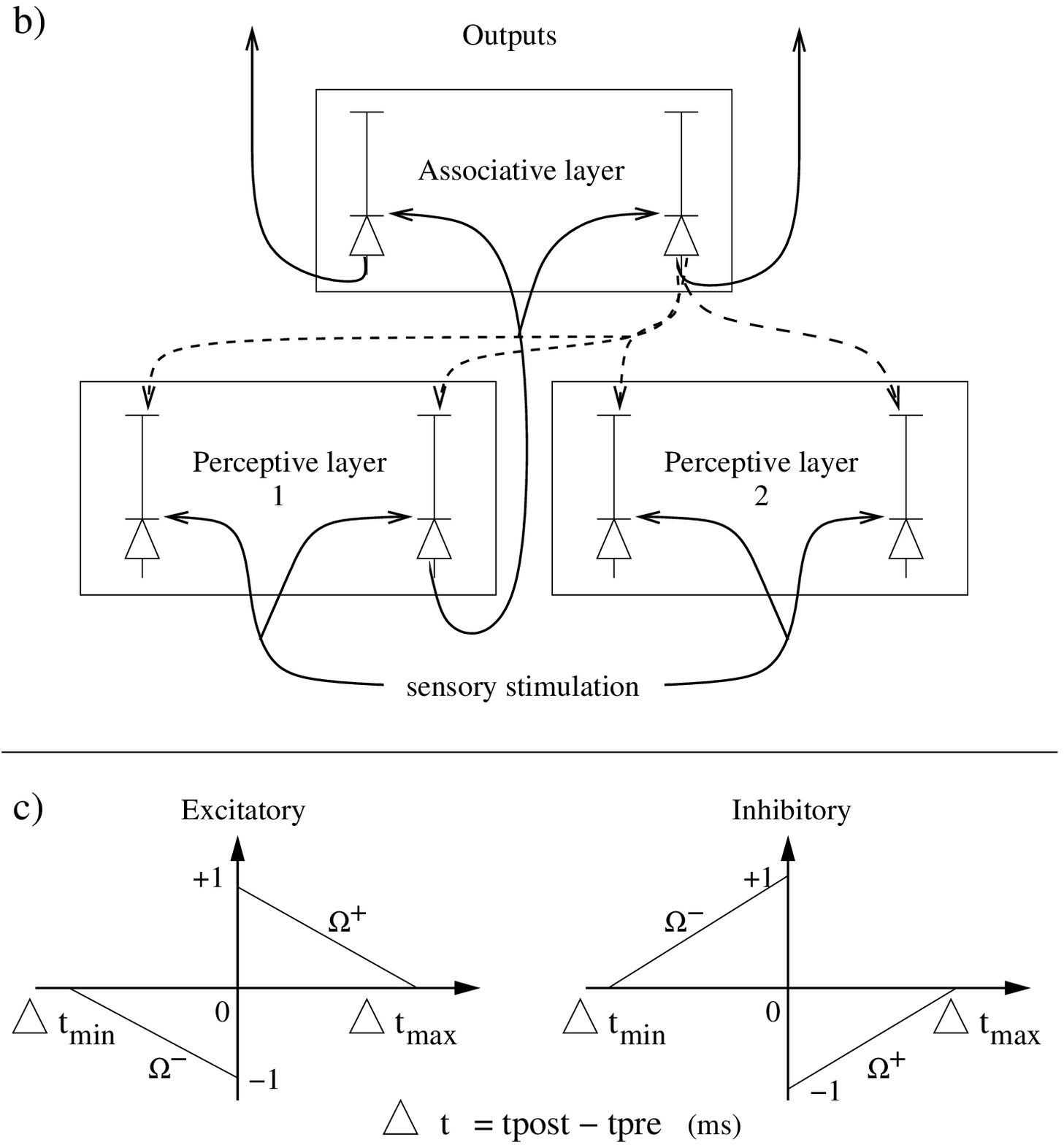}
\end{tabular}
\caption{\small a) Neocortical layer V pyramidal neuron model. 
b) Architecture of the network. Pyramidal neurons (triangles) have two distinct dentritic 
sites : basal (bottom horizontal line) and distal (upper horizontal line). 
c) STDP excitatory and inhibitory windows. Weight variation functions $\Omega^-$ for $\Delta t<0$ 
and $\Omega^+$ for $\Delta t>0$.}
\label{Fig:archi_neur_stdp}
\end{figure}

\paragraph{\bf Neuron model and network architecture}
Neurobiological studies have shown that neocortical layer V pyramidal neurons 
have specific abilities to integrate distal (layers I and II) top-down inputs and 
proximal (thalamic, layers V and VI) bottom-up inputs thanks to a dendritic 
and axonal action potential (AP) initiation sites \cite{LZ02,LZS01}. 
Hence we retain neocortical layer V pyramidal neurons 
as basic model of neuron (Figure~\ref{Fig:archi_neur_stdp}a) for our implementation 
of bimodal bidirectional associative memory.

Based on a classical BAM architecture (Bidirectional Associative Memory), adapted for multimodal 
association \cite{Rey05}, and also inspired from the architecture in \cite{SKK99}, 
we define a network (Figure~\ref{Fig:archi_neur_stdp}b) of three layers with 
100 excitatory and inhibitory neurons each. Sensory stimulations 
are received on the two perceptive layers, one for each modality. The 
associative layer plays a role of data fusion and gives the output pattern. 

\paragraph{\bf Neural dynamics and interconnections}
Excitatory neocortical layer V pyramidal neurons are coupled with inhibitory interneurons 
for dynamically modulate the amount of activity in the network (for safe of clarity, 
inhibitory interneurons are not represented on Figure~\ref{Fig:archi_neur_stdp}b).
The neurons are modeled in spike time coding, on the basis of the 
Spike Response Model \cite{GK02}. At each time $t$, the membrane potential $u_{j}$ of a postsynaptic neuron $j$ depends on the spike emission times $t_{i}^{(f)}$ of every presynaptic neurons $i$, weighted by $\omega_{ij}$, and on its own past emission times $t_{j}^{(f)}$, whith $t_{j}^{(f)}$ denoting the $f^{th}$ firing of neuron $j$, according to the equation

\begin{equation}
u_{j}(t)={\displaystyle \sum_{f}\eta(t-t_{j}^{(f)})}+{\displaystyle \sum_{i}}{\displaystyle \sum_{f}\omega_{ij}\varepsilon(t-t_{i}^{(f)})}
\label{potmemb}
\end{equation}
where $\eta(t)$ is the hyperpolarisation kernel of neuron $j$ and $\varepsilon(t)$ is the postsynaptic response kernel. For excitatory pyramidal neurons only, if top-down influences are activated, an extra term ${\sum_{Ca}\rho(t-t_{j}^{(Ca)})}$ is added to the righthand side of equation (\ref{potmemb}), where $\rho(t)$ is the response kernel to calcium action potentials CaAPs emitted at $t_{j}^{(Ca)}$ from the distal tufted dendrites emission site towards the somatic initiation site \cite{LZS01} (Figure \ref{Fig:archi_neur_stdp}a).

The emissions of CaAPs (spikes) are function of both the depolarization of distal dendritic site of neuron $j$ via backpropagated sodium action potentials NaAPs and distal dendritic excitation above a threshold $\Theta_{Ca}$ via pre-synaptic spikes \cite{LZ02}. The coincidence of backpropagated NaAPs and $\Theta_{Ca}$ crossing gives rise to CaAPs, thus facilitating a spike emission for neuron $j$.

The two perceptive layers (Figure \ref{Fig:archi_neur_stdp}b) receive sensory inputs on the basal dendritic site of the excitatory pyramidal neurons. Each pyramidal neuron makes a synapse to the basal site of the associative layer neurons, which in turn are connected to distal dendritic sites of the perceptive neurons, without layer distinction. Hence, perceptive pyramidal neurons compute the integration of top-down (associative) and bottom-up (sensory) inputs. At each pyramidal neuron in each layer we associate one strongly connected interneuron which makes inhibitory synapses with every other pyramidal neurons of the same layer. 


\paragraph{\bf Model of synaptic plasticity}

The weight $w_{ij}$ from a presynaptic neuron $i$ to a postsynaptic neuron $j$ is modified at each spike emission time $t_{pre}$ or $t_{post}$. Let $\Delta t = t_{post} - t_{pre}$ be the time difference between the last post and presynaptic spikes and $\Delta \omega_{ij}$ the weight variation of synapse $(i,j)$, subject to the bounds $0$ and $1$, or$-1$ for inhibitory synapses. We chose the following multiplicative rule for synaptic plasticity

\begin{equation}
\Delta\omega_{ij}=\left\{ 
\begin{array}{c}
(1-\omega_{ij})\times\Omega^{+}(\Delta t),\ \ \ \ \mbox{if}\ \Delta t>0\\
\omega_{ij}\times\Omega^{-}(\Delta t),\ \ \ \ \ \ \ \ \ \ \ \,\mbox{if}\ \Delta t<0
\end{array}
\right.
\label{stdp}
\end{equation}
where $\Omega^{-}$ and $\Omega^{+}$ represent the linear weight variation functions of Figure \ref{Fig:archi_neur_stdp}c.



\section{Bimodal association task}

\paragraph{\bf Bimodal stimulation and output}
A set of bimodal stimulations is a set $S$ of pairs $s_l = (s^1_l,s^2_l)$ where $s^k_l$ is the input vector of perceptive layer $k$. In the present article, the network has been tested on a set $S = \left\{s_1 ... s_{10}\right\}$ of $10$ pairs of characters represented by $10 \times 10$ matrices of black or white pixels. Associations to be learned are (`A',`1'), (`B',`2') and so on. An input pattern $s_l$ is presented repetitively $10$ times to the network, during $1ms$ every $10 ms$. Each $10 ms$ time slot, a black pixel $s^k_l(j)$ causes a spike emission of the pyramidal neuron $j$ on perceptive layer $k$. No spikes are caused by white pixels. The output $o_l$ of the network is the spiking activity pattern of the associative layer recorded along the $100 ms$ presentation of a bimodal stimulation $s_l$. Output can be represented by a matrix of $100$ (\# associative neurons) $\times 100$ (\# elementary $1ms$ time slots) binary values: $1$ for an output spike emission, $0$ otherwise.


\begin{figure}[hbt]
\centerline{\epsfig{file=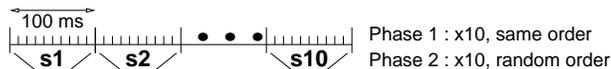, width=80mm}}
\caption{Stimulation protocol. Phase 1: Learning. Phase 2: Discrimination.}
\label{proto}
\end{figure}

\vspace{-6mm}
\paragraph{\bf Protocol}
The objective is to evaluate the ability of the model, first to reproduce a learned activity $k_l$ for each bimodal stimulation $s_l$, second to discriminate pairs of stimuli via different output activities $o_l$ for $l \in \{1 \dots 10\}$.
The experimental protocol consists in two successive phases of stimulation. Starting from random weights, in phase 1 the network adapts all the connections by STDP through $10$ repetitive presentations of a sequence of all the input patterns, in a given order (Figure~\ref{proto}). Phase~1 lasts $10s$ and is followed immediately by phase~2 where each pattern $s_l$ is again presented $10$ times to the network ($100 ms$ for each presentation), but the order of the $s_l$ is random and varies from one sequence to the next. All along the protocol, the weights of the network still continue varying through STDP learning. However phase~2 is mainly designed for testing the ability of the model to recall and discriminate all the patterns, independently of the sequence order imposed for learning stabilisation in phase~1. At each step of presentation of $S$, $o_l(p)$ is the output activity pattern of the network for stimulation $s_l$ at step $p$ ($p=1$ to $10$ in phase~1, $p=11$ to $20$ in phase~2). The learned activity $k_l = o_l(10)$ is defined as the activity pattern for $s_l$ at the end of phase~1. The same experimental protocol is applied in two conditions: A network without top-down modulations and a network with CaAPs top-down modulations.

\begin{figure}[hbt]
\includegraphics[scale=0.35]{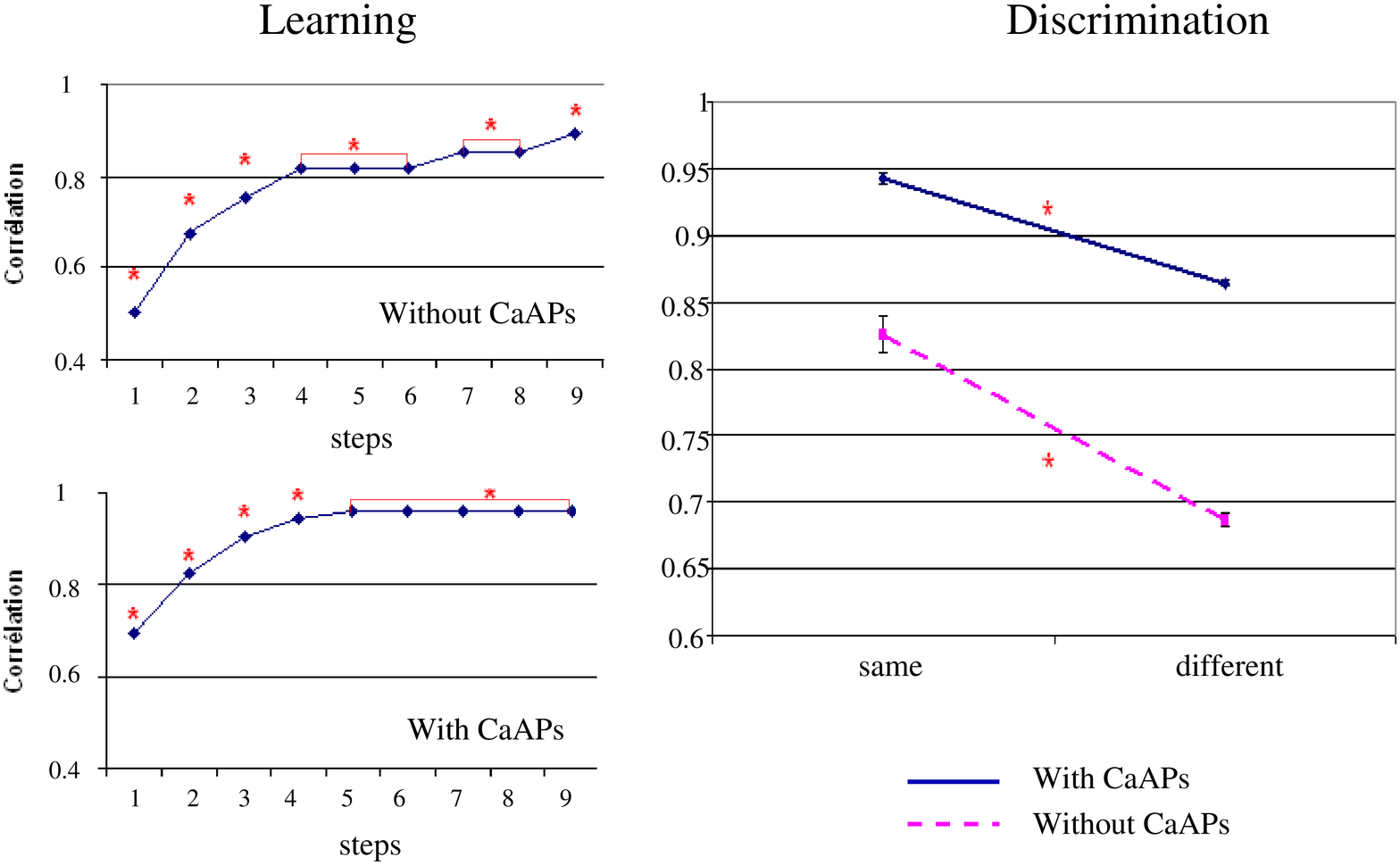}
\vspace{-4mm}
\caption{\small Learning and discrimination curves. Correlations between outputs in two conditions: Network without top-down modulations and network with CaAPs influences.}
\label{Fig:app_disc}
\end{figure}

\paragraph{\bf Analysis}
The matricial representations of the network outputs $o_l$ are compared by the method of {\em template correlation analysis}, as defined in \cite{LW01}. The method produces a template correlation coefficient, ranging between $-1$ and $+1$, which indicates the overlapping of two matrices, and then the similarity between two activity patterns. 
First we have studied, for all the stimulations $s_l$, the correlation of the successive outputs $o_l(p)_{1 \leq p < 10}$ in phase 1 to the learned activity $k_l$, in both conditions (Figure~\ref{Fig:app_disc} left). Second we have computed the average correlations between the learned activity $k_l$ and all the outputs $o_l(p)_{11 \leq p \leq 20}$ in phase 2, either for the same pattern, $k=l$, or for all the other stimulations, $k \neq l$ (Figure~\ref{Fig:app_disc} right, ``same'' or ``different'').



\section{Results and conclusion}

The experimental protocol has been applied to five networks with different initial weights. Figure~\ref{Fig:app_disc} presents the mean values of correlation coefficients, computed on all the stimulations and all the networks, for learning  (Student test) and discrimination (ANOVA), without and with top-down modulations. The values that are globally significant w.r.t the others are marked by a star.

Figure~\ref{Fig:app_disc} (left) shows that the correlation coefficients quickly increase in the first steps of learning phase~1 and reach values close to $0.9$. That proves a good stability of the network output for every stimulations learned by STDP, even in the absence of initial or supervised learning. The influence of top-down modulations is clearly positive: The outputs are better correlated with the learned activities, both in strength and time.



In discrimination (Figure~\ref{Fig:app_disc} right), the mean correlation coefficients measure the global ability of the model to discriminate patterns during the recall phase~2. Although the weights continue varying by STDP adaptation, learned activities are efficiently reproducible as specific outputs for the learned stimulations, even presented in random order. In both conditions, the correlation with learned activities is significantly higher for the same pattern (case {\em same}) than for all the other ones (case {\em different}). Top-down influences improve the correlation (close to $0.95$) but reduces the difference between the {\em same} and {\em different} cases.\\

The results show a stable activity obtained for each bimodal stimulation $s_i \in S$, reproducible during task and significantly different for each stimulation. We also show that top-down modulations can significantly increase the stability and reproducibility, with the consequence of reducing the differences between stimuli specific responses.
Our observations are coherent with the behaviour of general learning systems: High learnability power (e.g. VC-dimension) is not always suitable for good generalisation. However, in the framework of our experiments, our model of top-down modulated associative memory has both the abilities of fast learning ($4$ or $5$ steps of STDP are sufficient for convergence) and good discrimination (correlation with different learned activities is significantly lower than correlation with the stimulus specific response).

\begin{footnotesize}

\bibliographystyle{unsrt}
\bibliography{esann06_HAL}

\end{footnotesize}


\end{document}